\documentclass[10pt,twocolumn,letterpaper]{article}

\usepackage{3dv}
\usepackage{times}
\usepackage{epsfig}
\usepackage{graphicx}
\usepackage{amsmath}
\usepackage{amssymb}
\usepackage{booktabs}
\usepackage{caption}
\usepackage{subcaption}
\usepackage{enumitem}

\usepackage[pagebackref=true,breaklinks=true,colorlinks,bookmarks=false]{hyperref}
\usepackage[capitalize]{cleveref}

\threedvfinalcopy

\begin{document}

\title{MoCapDeform: Monocular 3D Human Motion Capture in Deformable Scenes}

\author{Zhi Li$^{1,2}$ $\quad$ Soshi Shimada$^{1,2}$ $\quad$ Bernt Schiele$^{2}$ $\quad$ Christian Theobalt$^{2}$ $\quad$ Vladislav Golyanik$^{2}$
\vspace{11pt}\\
$^{1}$Saarland University, SIC $\quad$ $^{2}$MPI for Informatics, SIC
} 

\twocolumn[{
\renewcommand\twocolumn[1][]{#1}
\maketitle
\begin{center}
    \centering
    \captionsetup{type=figure}
    \vspace{-10pt} 
    \includegraphics[width=\textwidth]{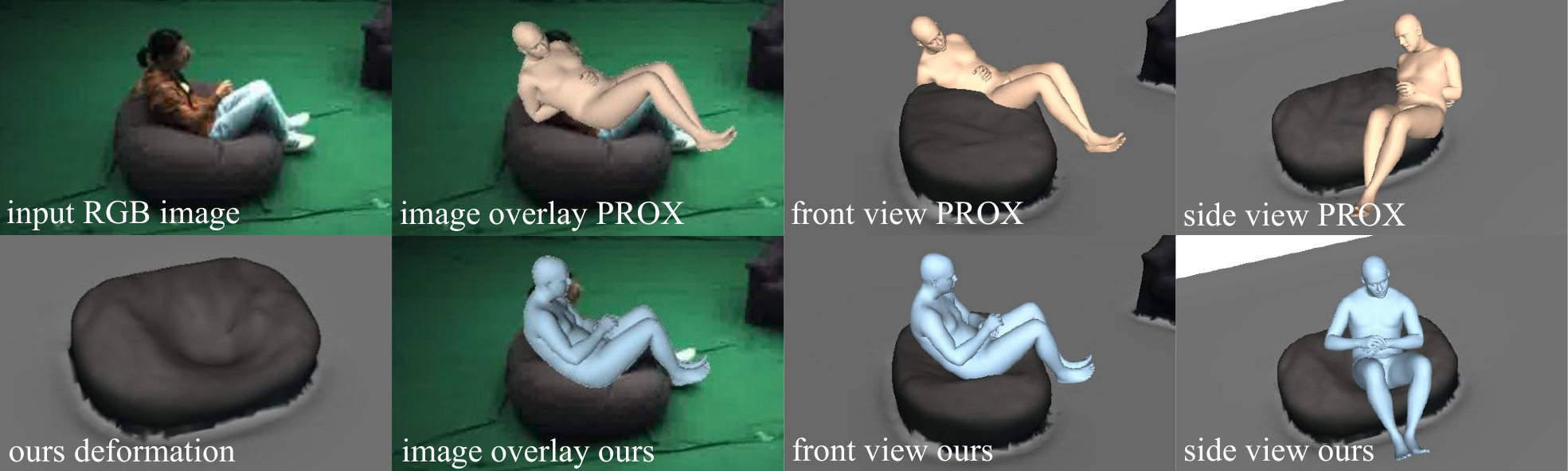}
    \captionof{figure}{Existing monocular 3D human motion capture methods such as PROX \cite{hassan2019resolving} ignore abundant scene deformation when penalising human-scene collisions, resulting in erroneous global poses (top). 
    \textbf{Our MoCapDeform algorithm is the first that models non-rigid scene deformations and finds the accurate global 3D poses of the subject by human-deformable scene interaction constraints}, achieving increased accuracy with significantly fewer penetrations (bottom).
    } 
    \label{fig:teaser}
\end{center}
}]

\begin{abstract} 
   3D human motion capture from monocular RGB images respecting interactions of a subject with complex and possibly deformable environments is a very challenging, ill-posed and under-explored problem. 
   Existing methods address it only weakly and do not model possible surface deformations often occurring when humans interact with scene surfaces. 
   In contrast, this paper proposes MoCapDeform,  \textit{i.e.,} a new framework for monocular 3D human motion capture that is the first to explicitly model non-rigid deformations of a 3D scene for improved 3D human pose estimation and deformable environment reconstruction. 
   MoCapDeform accepts a monocular RGB video and a 3D scene mesh aligned in the camera space. It first localises a subject in the input monocular video along with dense contact labels using a new raycasting based strategy. Next, our human-environment interaction constraints are leveraged to jointly optimise global 3D human poses and non-rigid surface deformations. MoCapDeform achieves superior accuracy than competing methods on several datasets, including our newly recorded one with deforming background scenes. 
\end{abstract}

\section{Introduction} 
\label{sec:intro} 
3D human motion capture from monocular images is an active research area 
\cite{martinez2017simple, pavlakos2018learning, tome2017lifting,  mehta2017vnect, rhodin2018unsupervised, kanazawa2018end,  dabral2019multi, shimada2020physcap, GraviCap2021, Wandt2022,  hakada2022unrealego}. 
Relying solely on monocular RGB inputs is challenging and severely ill-posed, as no explicit 3D cues are provided. 
As observed in daily life and noticed in the literature  \cite{gibson1950perception, hassan2019resolving}, the 3D world constrains human actions when they move and interact with it. 
Thus, environmental constraints and scene priors can provide additional cues for global 3D human motion capture. 
Leveraging them is a promising route to reduce the ambiguities and, at the same time, infer deformable scene geometry by observing human-scene interactions.

Existing works consider foot-floor interactions to recover physically plausible human motions \cite{shimada2020physcap,rempe2020contact,PhysAwareTOG2021,rempe2021humor}. 
Others utilise explicit 3D environment models as 
constraints \cite{hassan2019resolving,zhang2020perceiving,weng2021holistic,Zhang:ICCV:2021,guzov2021human}. 
These techniques either do not fully address the scale ambiguity \cite{shimada2020physcap,rempe2020contact,PhysAwareTOG2021,rempe2021humor,hassan2019resolving,zhang2020perceiving,weng2021holistic} or require RGB-D rather than RGB inputs \cite{Zhang:ICCV:2021}. 
Several other methods rely on 
body-mounted sensors (inertial measurement units) \cite{guzov2021human, PIPCVPR2022} to localise humans globally. 
Next, virtually all approaches consider the background 
being static and ignore potential scene changes caused by human-scene interactions. 
When a subject sits on a couch (lays in a bed), the latter  deforms significantly due to its non-rigidity and forces exerted by the subject. 
Unfortunately, while existing works  \cite{hassan2019resolving,zhang2020perceiving,weng2021holistic,Zhang:ICCV:2021,guzov2021human} use human-environment contact and inter-penetration constraints to avoid collisions, they simply disregard such scene deformations, which results in substantial 3D reconstruction errors.

We argue that 3D scene deformations 
cannot be ignored if we would like to raise the accuracy of 
the reconstructed 
3D human poses to the next level. 
Hence, this paper proposes \textit{MoCapDeform}, a new framework for 3D  monocular human motion capture with a 3D scene prior (a mesh); 
see Figs.~\ref{fig:teaser} and \ref{fig:framework} for an overview. 
In contrast to previous works, our method accurately localises the global human position in the scene from RGB inputs. 
It does so by a raycast contact finding policy to tackle the scale ambiguity problem, and a scene deformation modelling technique to address the limitation of low 3D reconstruction accuracy caused by the negligence of scene deformations. 

Our method comprises three stages. 
In the \textbf{first stage}, we initialise the 3D human poses parameterised by the SMPL-X model \cite{pavlakos2019expressive},  
which can be done by any off-the-shelf 3D human pose estimator. 
This yields initial root-relative 3D poses that are reasonably  accurate and sufficiently satisfy the image observations. 
Next, contact probability maps are estimated from the initial 3D poses, indicating which vertices on the human mesh are in contact with the 3D environment. 
The contact points are then re-projected to the image domain and passed through a raycasting operation to find the corresponding contact locations on the scene mesh. 
In the \textbf{second stage}, we register the estimated contact points on the human mesh to the raycasting results, leading to coarse global 3D human poses. 
\textbf{Finally}, these poses are further refined by jointly optimising for pose updates and deformations of the scene 
with which the body is in contact. 
In summary, our \textbf{contributions} are as follows: 
\begin{itemize}[noitemsep,nolistsep]
    \item MoCapDeform---the first framework for joint markerless 3D human motion capture from monocular RGB images and capture of non-rigid 3D scene deformations. 
    Such joint reasoning increases the accuracy of 3D human pose estimation on various benchmarks compared to existing methods (Sec.~\ref{sec:exp}). 
    \item A new raycasting based optimisation algorithm for finding dense contacts between humans and the environment  (Sec.~\ref{raycast}). 
    \item A joint scene deformation and human pose refinement optimisation to recover both accurate human poses and scene deformations (Sec.~\ref{deform}). 
    \item A new dataset for the experimental evaluation with  human-scene interactions and observable scene deformations (Sec.~\ref{ssec:datasets}). 
\end{itemize} 

We compare MoCapDeform to several previous state-of-the-art methods that assume monocular RGB images or videos and pre-scanned scene meshes input \cite{hassan2021populating, hassan2019resolving, pavlakos2019expressive}. 
Our approach regresses significantly more accurate global 3D human poses on the PROX  \cite{hassan2019resolving} and our new datasets, 
producing reasonable scene deformations (Sec.~\ref{sec:exp}). 
The new dataset and source code are available at  \url{https://github.com/Malefikus/MoCapDeform}.

\section{Related Work} 

\noindent\textbf{Kinematic Monocular 3D Human Pose Estimation.} 
Most works on monocular 3D human pose estimation reconstruct human joint positions in 
local coordinates. 
Some methods first estimate 2D poses in the 2D image space and then lift them into 3D  \cite{chen20173d,martinez2017simple,tome2017lifting,moreno20173d,fang2018learning,dabral2019multi}. 
Several other approaches learn feature representations for 2D poses  (without explicit 2D joint outputs) and perform lifting  \cite{mehta2017vnect,pavlakos2017coarse,habibie2019wild}. 
Further approaches regress 3D joints directly from RGB images via neural networks  \cite{tekin2016structured,mehta2017monocular,rhodin2018unsupervised}. 
While straightforward, this direct joint position representation has some issues, such as being hard to use in graphics applications, temporal jitter and implausible human  skeletons due to varying bone lengths. 
These limitations can be addressed by estimating parameters of pre-defined body models such as joint angles for kinematic skeletons \cite{zhou2016deep,mehta2017vnect,mehta2019xnect}, pose and shape parameters of parametric body models \cite{bogo2016keep,kanazawa2018end,pavlakos2018learning,pavlakos2019expressive,kocabas2020vibe, zhou2021monocular}, or template-based human performance capture methods  \cite{Habermann:2019:LRH:3313807.3311970, EventCap, deepcap, habermann21a}.
While most works estimate root-relative 3D human poses, 
several ones attempt to regress 3D human poses with absolute depths in the camera space  \cite{moon2019camera,shi2020motionet,mehta2019xnect}. 
Without depth priors, however, it is difficult to obtain accurate and  artefact-free human poses in the global reference frame.

\noindent\textbf{3D Human Pose Estimation with Scene Constraints.} While significant progress in 3D human pose estimation was made over the last decades  \cite{gavrila1999visual,moeslund2006survey,poppe2007vision, sarafianos20163d}, utilising scene constraints remains  insufficiently explored \cite{hassan2019resolving,  rempe2020contact, Zhang:ICCV:2021, rempe2021humor, GraviCap2021, ShimadaHULC2022}. 
Several works consider the ground plane only, and by enforcing volume occupancy exclusions or detecting foot-floor contacts, they impose physical plausibility on the reconstructed motions  \cite{zanfir2018monocular,shi2020motionet,rempe2020contact,shimada2020physcap,rempe2021humor}. 
Some works consider holistic representations and model complex scenes by placing arranged object meshes (recovered from categorised templates) into the desired coordinate frame  \cite{monszpart2019imapper,zhang2020perceiving,weng2021holistic}. 
This way, the knowledge about the spatial arrangement can be utilised to coarsely constrain the global position of the human. 
Another line of work employs pre-scanned meshes of the whole environment  \cite{hassan2019resolving,Zhang:ICCV:2021, wang2019geometric, cao2020long}. 
Hassan \textit{et al.}~\cite{hassan2019resolving} attempt to register empirically assumed contact vertices on the human body to the nearest scene vertices for human-scene collision detection. 
This approach does not fully resolve the depth ambiguity since the  exact contact locations in the scene are still unknown. 
Zhang \textit{et al.}~\cite{Zhang:ICCV:2021} use RGB-D videos as inputs and focus on the motion plausibility on top of scene constraints. 

All in all, the assumption of an available 
mesh enables accurately locating the human in the global reference frame, and provides the opportunity to model scene deformations for more accurate scene reconstruction and global human pose recovery. 
Considering the form of inputs (monocular RGB images + pre-scanned scene meshes) and outputs (global human poses), the PROX approach \cite{hassan2019resolving} is most related to our method. 
Another recent paper proposes HULC, \textit{i.e.,} a framework for 3D human motion capture in complex environments \cite{ShimadaHULC2022}. 
They estimate dense human-scene contacts with a neural network trained on a new large-scale dataset, and not a learning-free raytracing like we do. 
In contrast to previous and concurrent works  \cite{hassan2019resolving,Zhang:ICCV:2021,ShimadaHULC2022}, 
MoCapDeform models scene deformations, which helps to  accurately locate humans in the global 3D space.

\section{The Proposed Approach}

\begin{figure*}[t]
  \centering
  \includegraphics[width=\textwidth]{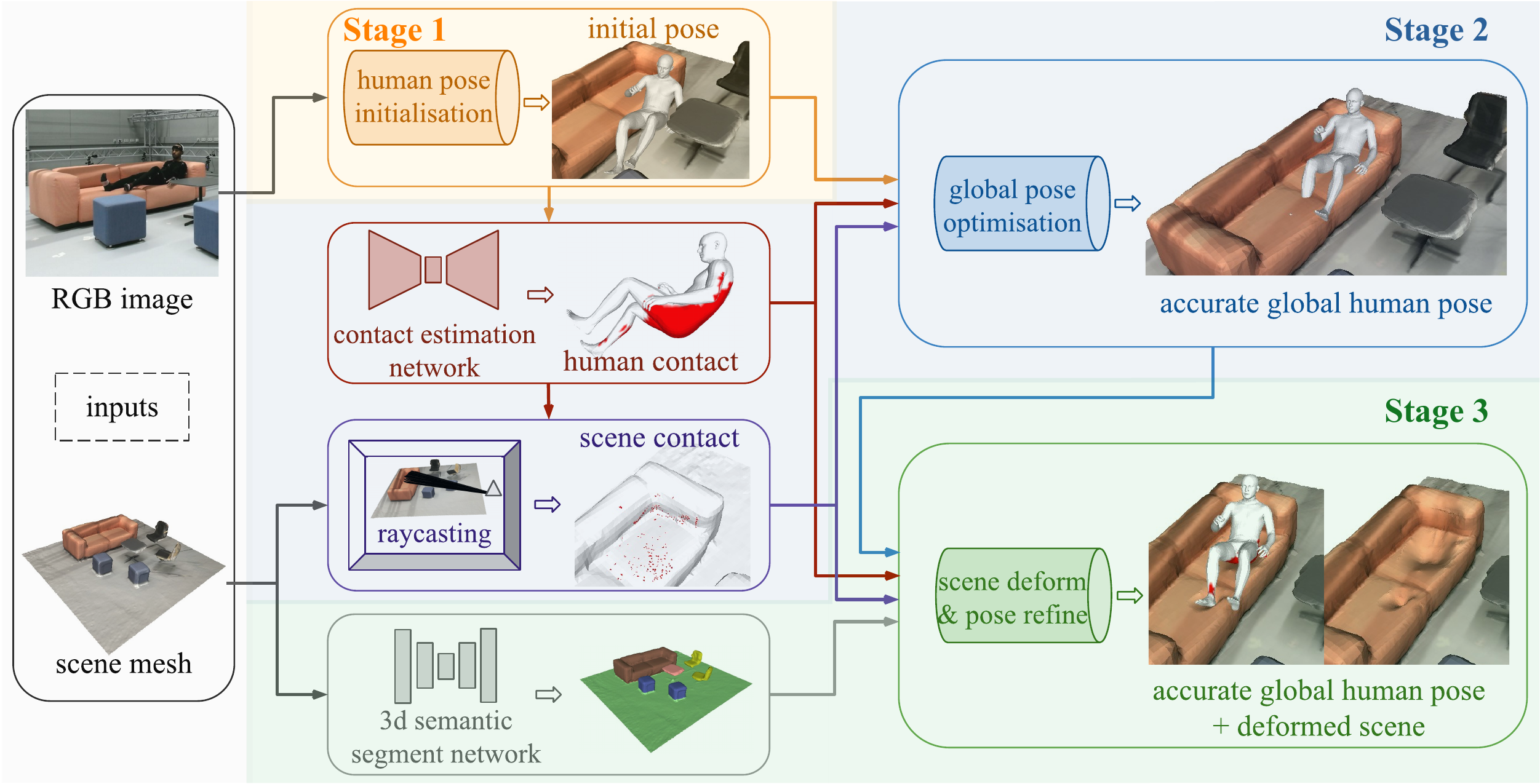}
   \caption{
   \textbf{Overview of MoCapDeform.}  
    We first initialise the human pose and 
    use it to find the contact points on the human mesh. 
    Then, we apply raycasting to find the contact points on the scene mesh surface, which are then used to recover improved global human poses. 
    Finally, we perform joint scene deformation and human pose refinement and obtain accurate global human pose and realistic scene deformations. 
   } 
   \label{fig:framework}
\end{figure*}

We describe our optimisation framework with three stages; see Fig.~\ref{fig:framework}.
The inputs to our framework are RGB images of a static camera and a pre-scanned mesh of the scene at the beginning of capture aligned to the camera coordinate frame; the outputs are posed 3D human bodies and deformed scene meshes. 
The \textbf{first stage} (Sec.~\ref{sec:stage1}) performs \textit{initial pose estimation}, where we estimate the initial pose from an RGB image, which suffers from scale ambiguity but faithfully overlays onto the images, \textit{i.e.,} the root-relative pose is coarsely accurate, on top of which the contact probability map can be estimated. 
The \textbf{second stage} (Sec.~\ref{raycast}) is a \textit{global pose optimisation}, in which we utilise the estimated contact points on the human mesh, then cast camera rays through the human contact points to the scene mesh to find the contact points on the environment, and then optimise for the global poses respecting these contacts. 
Finally, in \textbf{stage three} (Sec.~\ref{deform}), we perform \textit{joint scene deformation and pose refinement} to obtain accurate global 3D human poses and realistic scene deformations.

\subsection{Assumptions and Notations} 

Our method assumes that a 3D mesh of the scene and its registration in camera space are given. %
The 3D mesh is represented by $M_s=(\mathbf{V}_s,\mathbf{F}_s)$, with vertices $\mathbf{V}_s\in \mathbb{R}^{N_v \times 3}$ ($N_v$ as the number of vertices) and triangular faces $\mathbf{F}_s\in \mathbb{N}^{N_f \times 3}$ ($N_f$ as the number of faces) containing the vertex indices of each triangle. The static 3D scene mesh can be reconstructed with standard commercial solutions, either 
leveraging Structure Sensor \cite{structuresensor} and the Skanect \cite{skanect} software \cite{hassan2019resolving}, or multi-view reconstruction and differentiable rendering techniques 
from Agisoft Metashape \cite{Metashape}. The reconstructed meshes contain surface normals that correctly indicate the ``outside'' and the ``inside'' of the scene. Based on the topology of the pre-defined scene mesh, our method outputs a deformed per-frame
scene mesh  $M'_s=(\mathbf{V'}_s,\mathbf{F}_s)$ 
(we omit frame indices in this notation for conciseness).

Following the representation of \cite{hassan2019resolving}, %
we adopt a parametric 3D body model SMPL-X \cite{pavlakos2019expressive}. 
The model is formulated as a differentiable function $M_b(\boldsymbol{\beta,\theta,\psi,\gamma})$ parameterised by shape $\boldsymbol{\beta} \in \mathbb{R}^{10}$, body, hand and jaw pose $\boldsymbol{\theta}\in \mathbb{R}^{52\times 3}$ of axis-angle representation with three degrees of freedom (DoF) for each joint, holistic facial expression parameters $\boldsymbol{\psi} \in \mathbb{R}^{10}$, and global translation $\boldsymbol{\gamma}\in \mathbb{R}^{3}$ defining the global position of the pelvis joint. 
The model output is a 3D human mesh $M_b=(\mathbf{V_b,F_b})$, with vertices $\mathbf{V_b}\in \mathbb{R}^{10475 \times 3}$ and the corresponding triangular faces $\mathbf{F_b} \in \mathbb{N}^{20908 \times 3}$ expressing the mesh connectivity. From the mesh vertices, we can regress an underlying rigged skeleton $J(\boldsymbol{\beta})$ with $55$ joints 
defined by linear blend skinning. 
Following the notation of \cite{bogo2016keep}, we denote the posed joints as $R_{\boldsymbol{\theta \gamma}}(J(\boldsymbol{\beta})_i)$ for each joint $i$, where $R_{\boldsymbol{\theta \gamma}}$ denotes the kinematic tree defined by the pose parameter $\boldsymbol{\theta}$ and translation vector $\boldsymbol{\gamma}$. 
With this representation, we obtain a globally posed and shaped human body, \textit{i.e.,} both the skinned mesh and the underlying articulation. 

\subsection{Stage1: Initial Human Pose  Estimation}\label{sec:stage1} 
Stage 1 %
initialises the global human poses from monocular RGB images. 
This can be done by any of the off-the-shelf 3D human pose estimators \cite{bogo2016keep,mehta2017vnect,kanazawa2018end,mehta2019xnect,hassan2019resolving,kocabas2020vibe}, which take the RGB images as input, follow closely the image cue and produce the required root-relative poses. 
As MoCapDeform is not restricted to sequential inputs, we employ a single-frame method for the initialisation. 
For a fair comparison with the SOTA approaches  \cite{hassan2019resolving, hassan2021populating}, which are both based on the optimisation-based SMPLify-X \cite{pavlakos2019expressive}, 
we initialise with SMPLify-X. 
Stage 1 minimises the following objective function:
\begin{equation} \label{eq:stage1}
\begin{aligned}
    E_1(\boldsymbol{\beta,\theta,\gamma}) =
    E_J + \lambda_\theta E_\theta + \lambda_\alpha E_\alpha + \lambda_\beta E_\beta. 
\end{aligned}
\end{equation}
$E_J$ is the RGB data term, \textit{i.e.,} is a re-projection loss  seeking to minimise the robust weighted distance between 2D joints---estimated from the RGB image using OpenPose \cite{wei2016convolutional,cao2017realtime,simon2017hand}---and the 2D projection of the corresponding posed 3D joints of SMPL-X. 
We assume a perspective camera model. 
The re-projections are weighted by the detection confidence scores, and a robust Geman-McClure error function \cite{geman1987statistical} is applied on top to down-weight noisy detections. 
$E_\theta$ is the trained VAE-based body pose prior of VPoser \cite{pavlakos2019expressive}, which enforces natural human poses learned from a large 3D human pose corpus \cite{AMASS:ICCV:2019}. 
$E_\alpha=\sum_{i\in(\text{elbows,knees})}\exp(\boldsymbol{\theta_i})$ is a prior penalising extreme bending for elbows and knees. $E_\beta$ is an $\ell_2$-regulariser for human shape to penalise deviation from the neutral state. 
For a more detailed explanation, please refer to  \cite{pavlakos2019expressive, hassan2019resolving}.

\subsection{Stage 2: Global Pose Optimisation}\label{raycast} 
The goal of Stage 2 is to regress an accurate global 3D position of the human body estimated in Stage 1. 
Since monocular 3D human pose estimation is ill-posed without further priors, it is unlikely to obtain accurate results. 
Hence, we use the given scene mesh as a constraint to tackle depth ambiguity. 
We utilise the human-scene contact information by firstly estimating human body contacts on the initialised human body.   
We then perform raycasting of the re-projected human body contacts into the 3D scene to find the corresponding scene contacts and, finally, register the human body contact points to the scene contacts. 

\subsubsection{Raycast Contact Estimation}
\begin{figure}[t]
  \centering
  \includegraphics[width=0.47\textwidth]{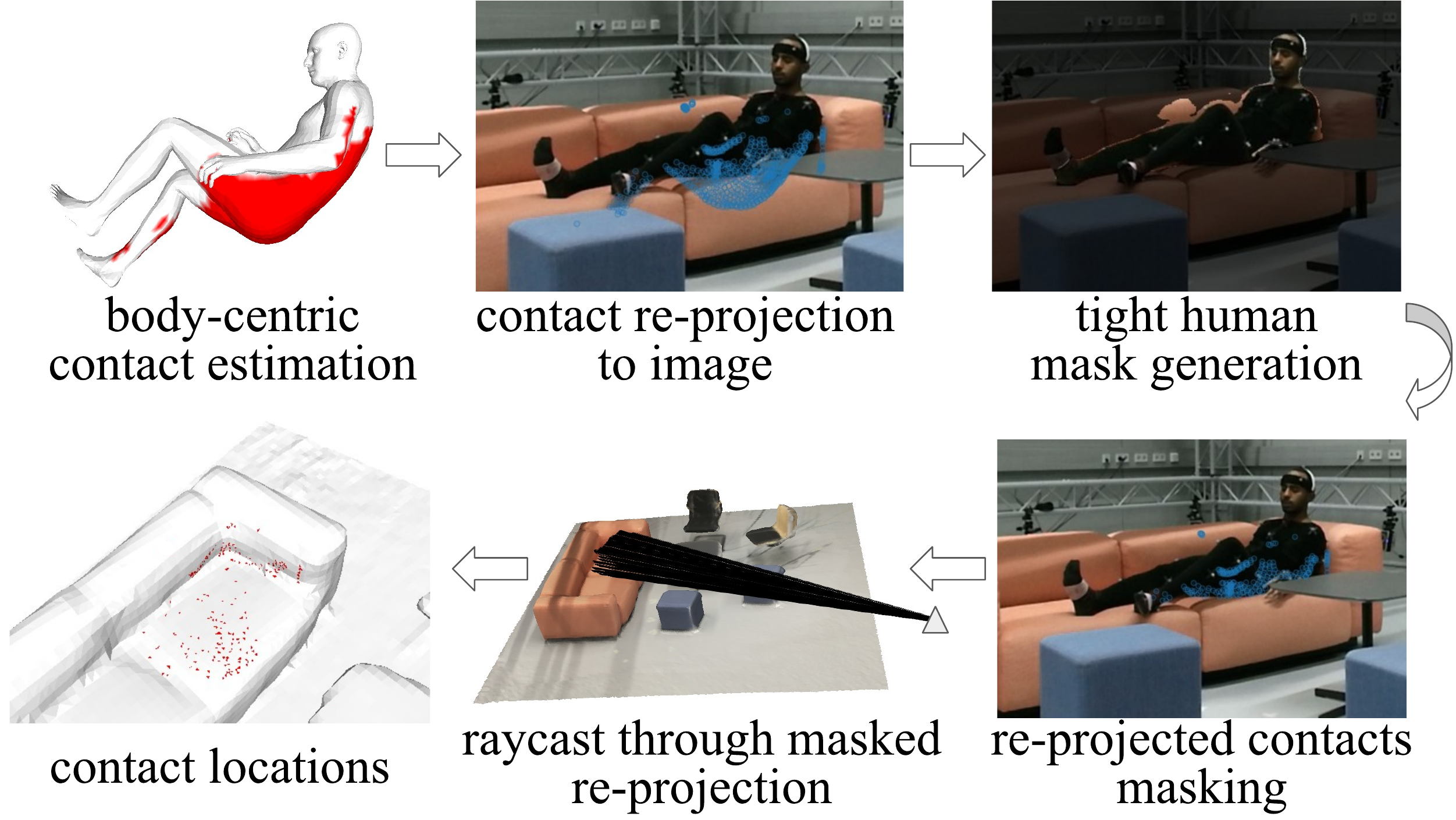}
   \caption{Overview of our raycast contact policy.} 
   \label{fig:raycast}
\end{figure}

Our raycasting policy has three steps illustrated in Fig.~\ref{fig:raycast}: 1) Body-centric contact estimation, 2) Contact re-projection with masking, and 3) Raycast and scene contact estimation. 

To find contacts on a scene through our raycast policy, we first need to find contacts on the human body. 
We thus employ POSA~\cite{hassan2021populating}, \textit{i.e.,} a conditional variational autoencoder (cVAE) %
that generates probability maps for different canonical human poses. 
The learned cVAE decoder takes as input %
human mesh points (in a root-relative pose and a canonical reference frame) 
and a latent vector $\boldsymbol{z}{\sim}\mathcal{N}(0,I)$ that conditions the sampling and can be directly applied to the initial human poses from Stage 1 since the root-relative poses are accurate. 
Specifically, we canonicalise the initialised human meshes from Stage 1 following the same formulation as in POSA and choose a zero vector  $\boldsymbol{z}$ 
to generate contact probability samples that 
lie in the peak in the learned latent space. 
Then, by thresholding the generated probability map by $0.5$ 
\cite{hassan2021populating}, 
we obtain the human mesh vertices that are in contact with the  environment. 

\begin{figure}[t] 
  \centering 
  \includegraphics[width=0.47\textwidth]{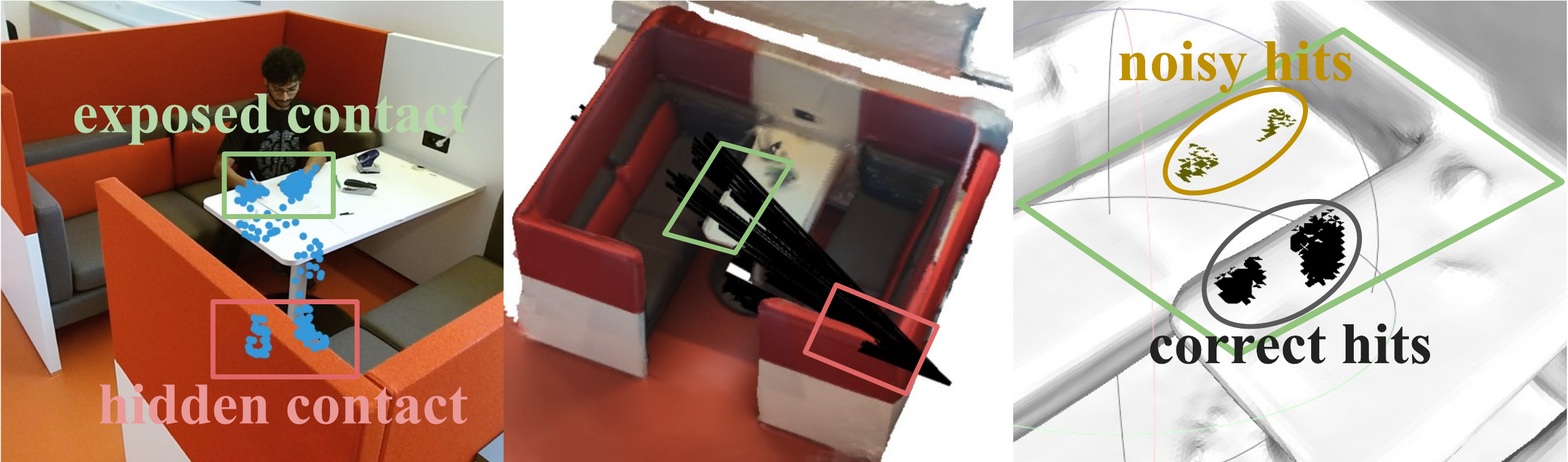}
   \caption{ 
   Detection of non-occluded areas and noisy contact label  filtering based on the analysis of the ray-mesh hits. 
   } 
   \label{fig:ray_hits} 
\end{figure}

The generated human contacts are then re-projected to the image  space. 
Then, we cast rays from the camera through these re-projections to
find intersections with the 3D scene mesh. 
However, as the scene geometry is complex, there are usually multiple hits along each ray, and the challenge is to find out which hit is at the contact. 
Intuitively, as illustrated in Fig.~\ref{fig:ray_hits}, when the re-projected contacts fall on the body parts that are not occluded in the image 
(the green rectangles in Fig.~\ref{fig:ray_hits}), 
the front-most hits will be at the correct contacts. 
Otherwise, the rays firstly hit occluders (red rectangles in Fig.~\ref{fig:ray_hits}). 
To tell apart the seen and occluded body parts, we apply %
PointRend~\cite{kirillov2020pointrend}, which finds tight human masks in the images. 
The masks are then used to segment out the contact re-projections on the occluded body parts. Furthermore, to eliminate the noisy raycast when the contact estimation is slightly off, we apply the DBSCAN clustering method \cite{ester1996density} for denoising, as shown in Fig.~\ref{fig:ray_hits}. 
We empirically set the scanning radius $\epsilon{=}0.5$ and MinPts${=}50$ of the DBSCAN, to help eliminate the clusters (hit by the ray) that are  far away  from the main cluster or contain a small number of samples. 
After these steps, the resulting points are  considered the corresponding scene contacts.

\subsubsection{Our Objective Function (Stage 2)} 
With the help of the raycast results, we perform Stage 2 optimisation and register the estimated human contacts to the raycasted scene contacts.  
This leads to refinement of the initial estimates from Stage 1. %
We minimise the following objective function: 
\begin{equation} \label{eq:stage2}
\begin{aligned}
\footnotesize
    E_2(\boldsymbol{\beta,\theta,\gamma},M_s)=E_J+\lambda_{\text{obs}}  E_{\text{obs}}+\lambda_{{\text{un}}}E_{\text{un}}
    +\lambda_{t\theta}E_{t\theta}+\lambda_{t\gamma}E_{t\gamma},
\end{aligned}
\end{equation}

\noindent where $E_J$ is %
as is in \eqref{eq:stage1}. 
$E_{\text{obs}}$ is the ``seen'' contact term, which minimises the distance between estimated contact points on the human mesh and the raycasted contact points on the scene mesh. 
$E_{\text{un}}$ denotes the ``unseen'' contact term, which registers the rest of the estimated human contacts 
that do not have raycast hits to the corresponding nearest scene vertices by minimising their Chamfer distance. 
In $E_{\text{obs}}$ and $E_{\text{un}}$, the distances are calculated by a Geman-McClure error function \cite{geman1987statistical} to downweight noisy detections. 
Finally, %
we apply temporal smoothness terms $E_{t\theta}$ and $E_{t\gamma}$. 
$E_{t\theta}$ and $E_{t\gamma}$ are the $\ell_2$-distances  between the poses $\boldsymbol{\theta}$ and global translations $\boldsymbol{\gamma}$ between frames $t$ and $t{-}1$, respectively.

\subsection{Joint Scene Deformation and Pose Refinement}\label{deform} 
In Stage 3, the coarse global poses obtained from Stage 2 are further refined, taking into account scene deformations. 
This stage jointly optimises for plausible 3D scene deformation in interaction regions and more accurate human poses, also resulting in fewer inter-penetrations between the two. 

\subsubsection{Scene Deformation Modelling}\label{scene-model}
To model 
deformations of 
non-rigid objects such as couches or beds, we use an  as-rigid-as-possible (ARAP) regulariser \cite{sorkine2007rigid}. 
It allows deforming a mesh with guidance by pre-defined sparse control vertices $c \in \Omega$ ($\Omega$ is a subset of contact point indices). 
The latter are first moved to the target positions, and the neighbouring faces are encouraged to stay rigid as much as possible. 
The deformed state of the entire surface is then found by optimising
\begin{equation} \label{eq:arap}
\footnotesize
    E_{\text{ARAP}}(M'_s, \mathbf{R}_c) = \sum_{c \in \Omega}\sum_{n \in \mathcal{N}(c)}w_{cn}\left \| (\mathbf{v'}_c-\mathbf{v'}_n)-\mathbf{R}_c(\mathbf{v}_c-\mathbf{v}_n) \right \|,
\end{equation}
\noindent where $\mathbf{R}_c$ is the unknown rotation matrices; $\mathbf{v}_c$ and $\mathbf{v'}_c$ are the control vertex  positions before and after the optimisation; $\mathcal{N}_c$ is the set of per-vertex neighbours  $\mathbf{v_c}$; $\mathbf{v}_n$ and $\mathbf{v'}_n$ are the neighbouring vertices of $\mathbf{v}_c$ before and after optimisation, and $w_{cn}$ are cotangent weights. 
To integrate ARAP regulariser in our framework, 
we define on the scene mesh: 1) A set of control vertices $\mathbf{v_c}$; 
2) The corresponding target positions $\mathbf{v'}_c$ for the control vertices; 3) A set of neighbouring vertices $\mathbf{v_n}$, which are allowed to move. 

\begin{figure}[t]
  \centering
  \includegraphics[width=0.47\textwidth]{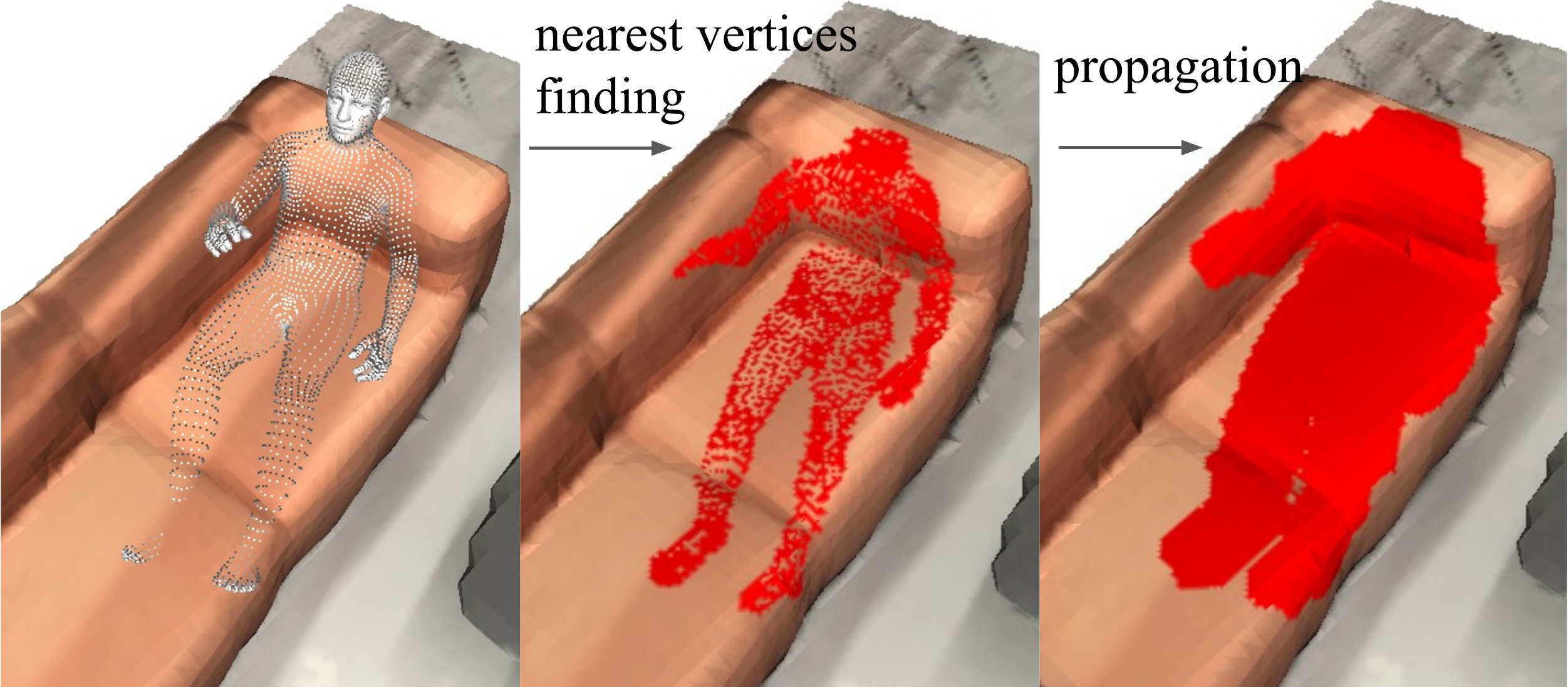}
   \caption{Determination of movable scene points. 
   }
   \label{fig:partition}
\end{figure}

In practice, with the help of the current state of the human body, we first partition the whole scene  mesh into movable and static areas and then choose  the control points from the movable area and define the target positions accordingly. 
In the beginning, we need to know which parts of the  scene are deformable and which are not. 
For that purpose, we adopt a 3D scene mesh segmentation network VMNet \cite{hu2021vmnet} trained on a large-scale indoor dataset ScanNet \cite{dai2017scannet}. 
VMNet estimates semantic labels of the furniture. 
In this paper, we regard ``sofa'' and ``bed'' as non-rigid and all other  object classes as rigid. 
After masking out rigid parts, %
we further define ``movable'' areas inside the deformable areas. 
As illustrated in Fig.~\ref{fig:partition}, for every vertex on the human mesh, we find its nearest scene vertex and then propagate the obtained points to their $k$-th order neighbours. 
These points are regarded as ``movable'' during ARAP deformation. 
Here, we empirically set $k{=}3$ according to the scale of the meshes. 

\begin{figure}[t]
  \centering
  \begin{subfigure}{0.49\linewidth}
    \includegraphics[scale=0.235]{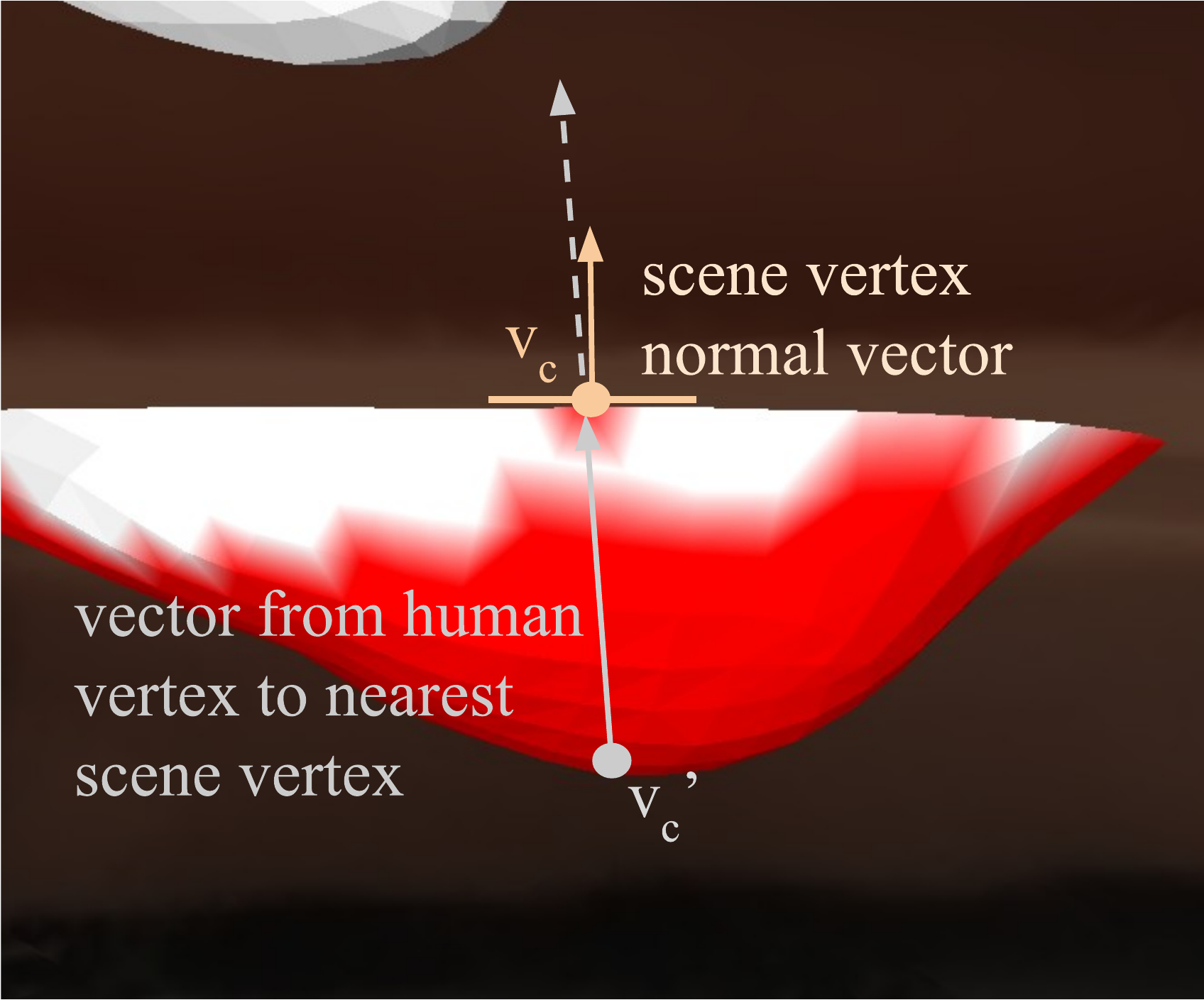}
    \caption{collision checking}
    \label{fig:normal-a}
  \end{subfigure}
  \hfill
  \begin{subfigure}{0.49\linewidth}
    \includegraphics[scale=0.235]{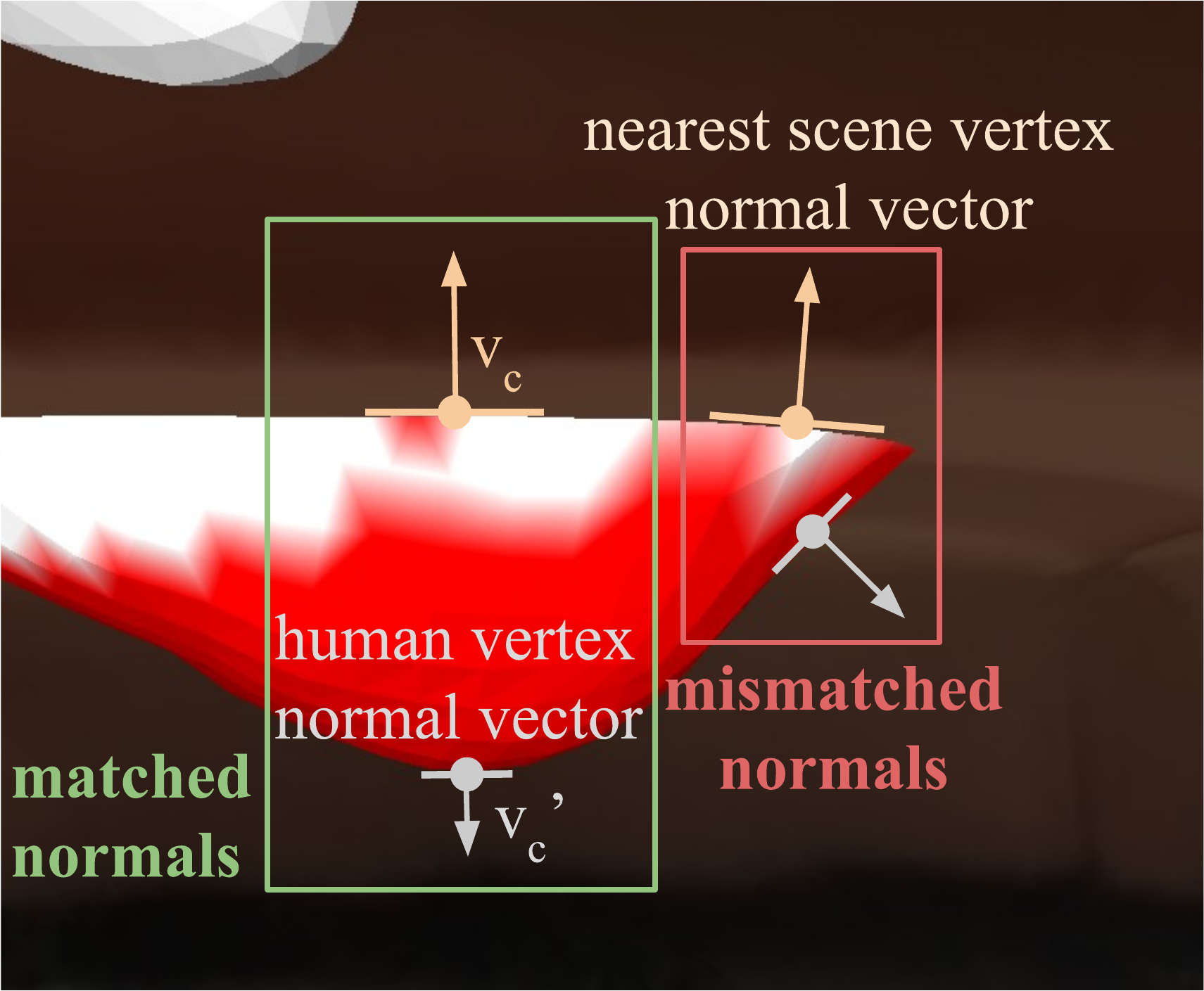}
    \caption{normal matching}
    \label{fig:normal-b}
  \end{subfigure}
  \caption{%
  Collision checking and normal matching. The viewpoint is ``inside'' the couch, looking at the colliding hip.
  }
  \label{fig:normal}
\end{figure}

Next, we define the control vertices $\mathbf{v}_c$ on the scene mesh and their corresponding target positions $\mathbf{v'}_c$. 
Assuming that the human induces the scene deformations, $\mathbf{v}_c$ and $\mathbf{v'}_c$ can be defined  according to the vertices $\mathbf{v}_{c'}$ of the  human mesh (and their positions) that satisfy the  following conditions: 
\begin{itemize}[noitemsep,nolistsep]
    \item $\mathbf{v}_c$ is the nearest scene vertex to the human vertex $\mathbf{v}_{c'}$.
    \item $\mathbf{v}_{c'}$ are marked as ``in contact'' by the contact estimation step.
    \item $\mathbf{v}_{c'}$ collide with the scene mesh. Note that in this step, collision detection cannot be achieved by the pre-computed SDF since the scene is deforming. 
    Hence, we check the collision status with the help of scene surface normals, as shown in Fig.~\ref{fig:normal-a}: When the vector from $\mathbf{v}_{c'}$ to $\mathbf{v}_c$ (grey) goes in the same direction as the normal vector of $\mathbf{v}_c$ (orange), $\mathbf{v}_{c'}$ will be classified as colliding with the scene.
    \item The normals of $\mathbf{v}_{c'}$ should be of opposite directions to the normals of $\mathbf{v}_c$, as is shown in Fig.~\ref{fig:normal-b}. This is because the scene usually deforms along the direction of the forces applied by the human. 
\end{itemize} 
At last, the nearest scene vertices $\mathbf{v}_c$ to the above human vertices $\mathbf{v}_{c'}$ are chosen as control points, and the control target positions $\mathbf{v'}_{c}$ are defined by the positions of $\mathbf{v}_{c'}$.

\subsubsection{Optimisation (Stage 3)} 
The final stage is a joint and alternating  optimisation for scene deformation and human poses. 
In every iteration $k$, we firstly pick the control points of ARAP according to the current human pose using the techniques described in Sec.~\ref{scene-model} and then update the scene mesh using  \eqref{eq:arap}. 
Next, we update the human pose based on the updated scene mesh by minimising the following energy function: 
\begin{equation} \label{eq:stage3}
\begin{aligned}
    E_3(\boldsymbol{\beta,\theta,\gamma},M_s^k) = E_2(\boldsymbol{\beta,\theta,\gamma},M_s^k) + \lambda_{\text{pen}}E_{\text{pen}}. 
\end{aligned}
\end{equation}
\noindent %
where $E_2(\boldsymbol{\beta,\theta,\gamma},M_s^k)$ is defined in
\eqref{eq:stage2} and 
$E_{\text{pen}}$ is a penetration term that does not use the pre-computed SDF, since the scene mesh is being constantly updated. 
Instead, it utilises the normal checking technique presented in Fig.~\ref{fig:normal-a} to detect collisions and then registers the colliding vertices on the human mesh to their nearest scene mesh vertices by minimising the Geman-McClure error function  \cite{geman1987statistical}.

\subsection{Implementation}
In Stage 1, closely following \cite{pavlakos2019expressive, hassan2019resolving}, we optimise \eqref{eq:stage1} using a PyTorch \cite{paszke2017automatic} implementation of the limited-memory BFGS optimiser %
\cite{nocedal2006nonlinear} with line search satisfying strong Wolfe conditions. 
\eqref{eq:stage2} in Stage 2 and \eqref{eq:stage3} in Stage 3 are optimised by PyTorch implementations of the Adam optimiser \cite{kingma2015adam}. 
For the scene deformations, \textit{i.e.,} \eqref{eq:arap} in  Stage 3, we adopt the ARAP implementation from Open3D  \cite{Zhou2018open3d}. 
The off-the-shelf components we adopt (\textit{i.e.,} PointRend \cite{kirillov2020pointrend}, VMNet \cite{hu2021vmnet} and POSA human contact estimation \cite{hassan2019resolving}) 
are easily deployable and sufficiently accurate for our task. 
The $\lambda$ and $w_{cn}$ weights in $\eqref{eq:stage1}$-$\eqref{eq:stage3}$ are empirically found and fixed in all experiments; see the source code. 

\section{Experiments}\label{sec:exp}
To evaluate our MoCapDeform framework, we conduct extensive experiments on two datasets  (Secs.~\ref{ssec:datasets}-\ref{sec:exp_quant}) and show qualitative results (Sec.~\ref{sec:exp_quali}). 

\subsection{Datasets}\label{ssec:datasets} 
\noindent \textbf{PROX dataset \cite{hassan2019resolving}.} The PROX dataset includes a large qualitative and a small quantitative sets.
The qualitative set contains monocular videos of $20$ human subjects interacting with $12$ indoor scenes along with the 3D scene scans: altogether, $100k$ RGB-D frames recorded at $30$ fps (without ground-truth 3D human poses). 
The quantitative set contains $180$ static RGB-D frames, with one human subject wearing markers interacting with a mimicked living room containing daily furniture. 
The pseudo-ground-truth SMPL-X parameters for the quantitative set are fitted by the marker-based MoSh++ \cite{AMASS:ICCV:2019} method. 

\noindent \textbf{MoCapDeform (MCD) dataset.} 
To evaluate all outputs of MoCapDeform, including the deformations, we record a new dataset of people interacting with furniture, \textit{i.e.,}
a non-rigid sofa, a deformable stool, and especially a beanbag, 
which retains its deformed shape after the interaction and 
allows obtaining ground-truth deformations. 
We reconstruct accurate human meshes and scene geometry with the multi-view camera setting and a markerless differentiable-rendering-based technique  \cite{Metashape}. 
The human meshes can then be used to fit the SMPL-X model parameters and serve as ground truth. 
The dataset contains four video sequences at $30$ fps of four subjects  interacting with the furniture ($16k$ sequential RGB images in total). 
We utilise the dataset for both quantitative and qualitative  experiments. 

\begin{figure*}[t]
  \centering
  \includegraphics[width=\textwidth]{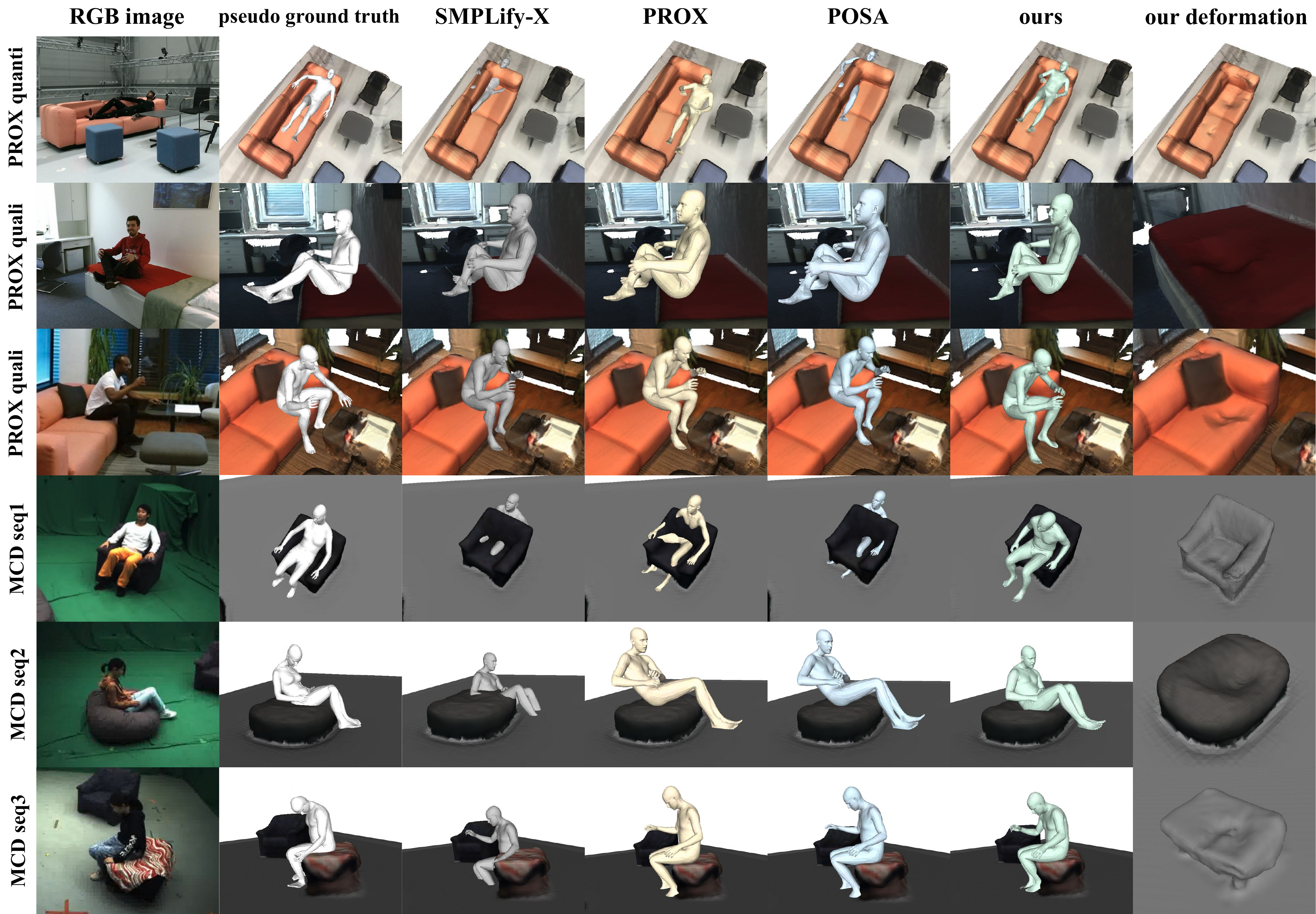}
   \caption{Qualitative results and comparisons on different datasets. Our MoCapDeform achieves more accurate global localisations than the state-of-the-arts, leads to less penetration, and prevents the human bodies from floating when there are large scene deformations. Moreover, it outputs plausible scene deformations not reconstructed by the previous methods. 
   } 
   \label{fig:quali}
\end{figure*}

\subsection{Quantitative Evaluation}\label{sec:exp_quant}

\begin{table}[t]
\centering
\footnotesize
\scalebox{0.92}{
\begin{tabular}{lccccc}
    \toprule
    & PJE & V2V & p.PJE & p.V2V & non-col\\
    \midrule
    SMPLify-X \cite{pavlakos2019expressive}   & 214.64    & 219.20    & 64.04   & 61.88  & 92.42\%\\
    PROX \cite{hassan2019resolving}    &  167.16   & 171.35    & 63.54   & 63.06   & 95.63\%\\
    POSA \cite{hassan2021populating}    & 157.11    & 159.52    & 63.70    & 63.23   & 95.89\%\\
    \specialrule{0.05em}{1pt}{1pt}
    \textbf{MoCapDeform (s1+s2)}    & \textbf{\textit{144.15}}    & \textbf{\textit{145.23}}    & \textbf{\textit{62.86}}    & \textbf{\textit{61.19}}  & \textbf{\textit{95.90\%}}\\
    \textbf{MoCapDeform (full)}    & \textbf{139.78}    & \textbf{140.60}    & \textbf{62.29}    & \textbf{60.67}  &\textbf{97.60\%}\\
    \bottomrule
\end{tabular}
}
\caption{Results on the PROX dataset using RGB inputs. We show our results of Stages 1 and 2 (``s1+s2'') and full method and compare them with several state-of-the-arts. 
Best is indicated in \textbf{bold}, and second best in \textbf{\textit{bold italic}}.
} 
  \label{tab:proxq}
\end{table}

\begin{table}[t]
\centering
\footnotesize
\scalebox{0.9}{
\begin{tabular}{lccccc}
    \toprule
    & PJE & V2V & p.PJE & p.V2V & non-col\\
    \midrule
    SMPLify-X \cite{pavlakos2019expressive}   & 441.86    & 451.87    & \textbf{89.73}   & \textbf{101.53}  & 97.14\%\\
    PROX \cite{hassan2019resolving}    &  375.01   & 403.22    & 97.09   & 107.57   & 97.99\%\\
    POSA \cite{hassan2021populating}    &  365.91   & 398.15 &   97.26  &   108.67  & 98.41\%\\ 
    \specialrule{0.05em}{1pt}{1pt}
    \textbf{MoCapDeform (s1+s2)}  &  \textbf{\textit{266.18}}  &  \textbf{\textit{283.46}}   &  \textbf{\textit{91.18}}   &   \textbf{\textit{101.71}} &  \textbf{\textit{98.57\%}}\\
   \textbf{MoCapDeform (full)}   &  \textbf{264.68}   &  \textbf{282.01}   &  91.91  &  102.43 & \textbf{99.04\%}\\
    \bottomrule
\end{tabular}
}
\caption{Results on MoCapDeform dataset using RGB inputs. 
We compare outputs of Stages 1 and 2 (``s1+s2'') and our full method 
to several state-of-the-art approaches. 
Best is indicated in \textbf{bold}, and second best in \textbf{\textit{bold italic}}.}
  \label{tab:selfrec}
\end{table}

\begin{table}[t]
\centering
\footnotesize
\scalebox{0.95}{
\begin{tabular}{lccccc}
    \toprule
    & PJE & V2V & p.PJE & p.V2V & non-col \\
    \midrule
    SMPLify-D \cite{hassan2019resolving}   & 70.63    & 72.19    & 44.58   & 44.33   & 93.65\%\\
    PROX-D \cite{hassan2019resolving}    &   63.03  & 65.64    & 39.89   & 39.74   & 93.86\%\\
    POSA-D \cite{hassan2021populating}    &   62.44  & 66.16    & 39.73   & 40.11   & 93.97\%\\
    \specialrule{0.05em}{1pt}{1pt}
    \textbf{MoCapDeform (s1+s3)} & \textbf{59.32}   & \textbf{62.37}   & \textbf{39.57}  & \textbf{39.12} & \textbf{97.04\%}  \\
    \bottomrule
\end{tabular}
}
\caption{Results on the PROX quantitative dataset using RGB-D inputs. Best is indicated in \textbf{bold}.}
  \label{tab:proxd}
\end{table}

We evaluate the estimated 3D human poses by computing several quantitative metrics indicating the global and local 3D reconstruction accuracy and the degree of penetrations. 
For global 3D reconstruction accuracy, we adopt the standard \textbf{PJE} and \textbf{V2V} metrics and report them in $mm$. 
PJE stands for the mean per joint position error, and V2V indicates the mean vertex-to-vertex error. 
For local 3D reconstruction accuracy, we employ \textbf{p.PJE} and \textbf{p.V2V} (in $mm$), which are the PJE and V2V metrics after Procrustes alignment. 
Furthermore, to evaluate the human-scene penetrations---following the work in this domain \cite{hassan2019resolving,zhang2020place,zhang2020generating,Zhang:ICCV:2021}---we report the \textbf{non-collision score}, which is the percentage of human body mesh vertices that do not penetrate the scene mesh. 
Note that for MoCapDeform, the non-collision score is calculated over the  deformed  meshes since they are also an output of our method.

The results on the PROX dataset and our new dataset are summarised in Tables \ref{tab:proxq} and \ref{tab:selfrec}. We report the results of Stages 1+2 and all stages of our framework (full model). 
For our new MoCapDeform dataset, we down-sample the  framerate to $5$ fps. 
As can be observed in the tables, both the global pose optimisation and the scene deformation stages contribute to more accurate pose estimation and outperform the previous approaches. 
Our method achieves significant improvement in terms of the global poses. 
With the help of the estimated scene deformations, the final  output meshes of MoCapDeform have significantly fewer  penetrations. 

To further evaluate the effectiveness of the scene deformation stage, we conduct experiments on top of the RGB-D inputs from the PROX quantitative dataset; see Table \ref{tab:proxd}. 
Specifically, we replace the pose initialisation stage with the PROX-D method \cite{hassan2019resolving}, in which a depth term is used as a constraint during optimisation, supposedly resolving the depth ambiguity. 
Then we skip the second stage and directly apply our joint scene deformation and pose refinement stage over the PROX-D initialisation, with a depth data term (the same one as used in \cite{hassan2019resolving}) added to \eqref{eq:stage3}. 
The numbers in Table \ref{tab:proxd} show that our scene deformation stage can further improve the accuracy of 3D human pose estimation in all metrics and results in significantly fewer inter-penetrations after the deformation (\textit{cf.}~non-collision score). 

\subsection{Qualitative Results}\label{sec:exp_quali}

\begin{figure}[t]
  \centering
  \includegraphics[width=0.5\textwidth]{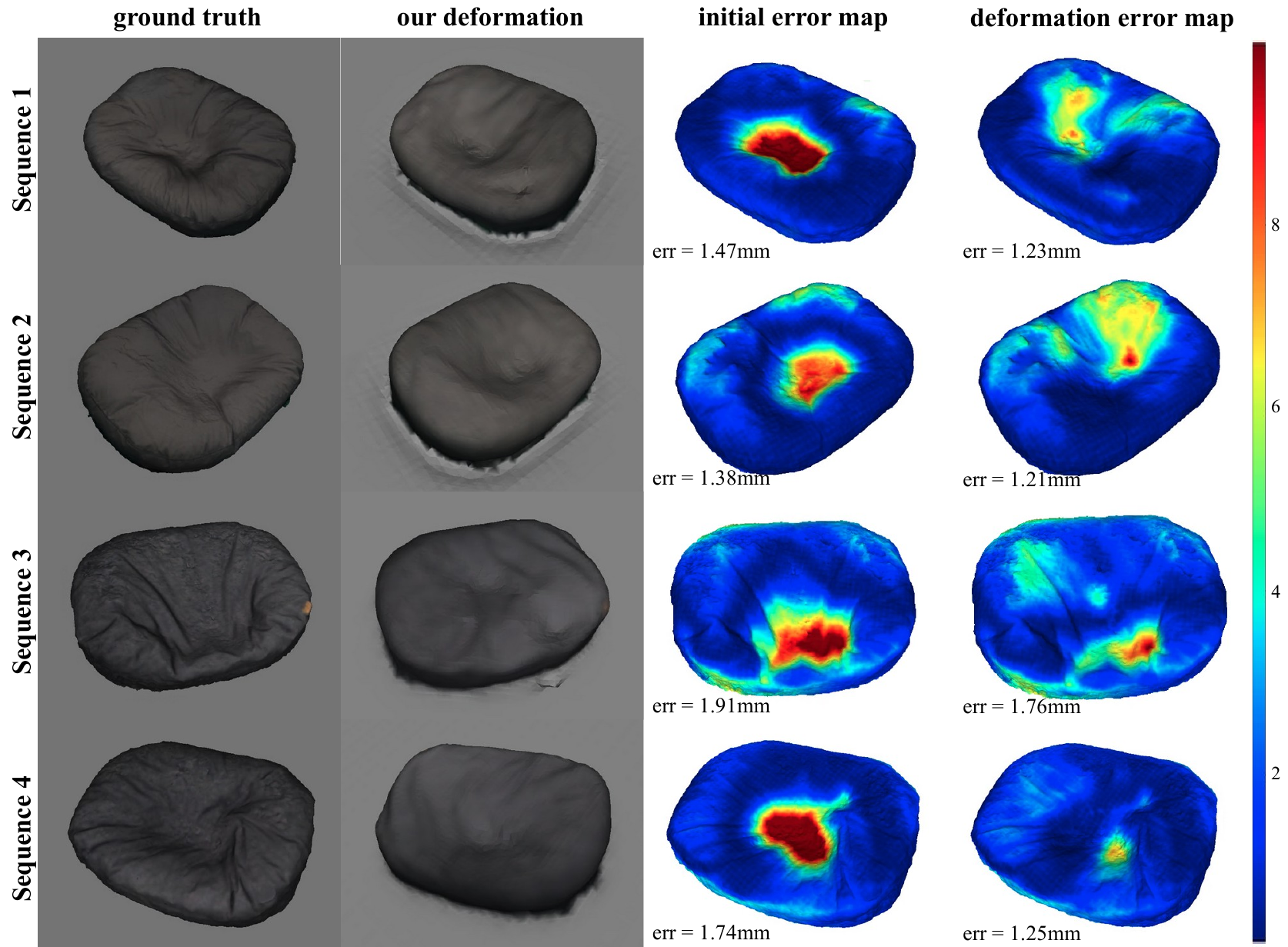}
   \caption{Comparison of ground-truth meshes and our deformations. 
   The error maps show colour-coded per-vertex distances between the ground-truth meshes and the
   initial shapes or final states estimated by MoCapDeform. 
   }
   \label{fig:deform}
\end{figure}

We show qualitative results on PROX and our MoCapDeform datasets in  Fig.~\ref{fig:quali}. 
SMPLify-X \cite{pavlakos2019expressive} inaccurately localises the human and causes 
severe penetrations, as it ignores scene information. 
Both PROX \cite{hassan2019resolving} and POSA \cite{hassan2021populating} leverage scene constraints by penalising the human-scene collisions with pre-computed SDF values of the static scenes. 
Since they do not model scene deformations, the collision  penalising terms tend to lift the human above the scene surfaces even in the presence of large scene deformations. 
This leads to the subject floating 
(Fig.~\ref{fig:quali}, PROX: rows 1, 2 and 5; POSA: rows 2 and 5), causing inaccurate global positions along the depth channel or severe body-scene penetrations (Fig.~\ref{fig:quali}, PROX: rows 3 and 4; POSA: rows 1, 3 and 4), as the image cues and  anti-collision terms cannot be satisfied  simultaneously. 
In contrast, with the help of our raycast contact algorithm and scene deformation modelling, MoCapDeform finds more accurate global human positions without the floating issue, with  significantly fewer inter-penetrations, and outputs plausible  scene deformations. 
See Fig.~\ref{fig:deform} for 
comparisons 
between the ground-truth states of the beanbag (reconstructed after the person stands up) and the deformation output from our method.

\section{Discussion and Conclusion} 
One limitation of our method is the dependency on reasonable 3D pose initialisation; recovery from starkly erroneous initial poses is unlikely. Moreover, severe scene occlusions blocking all human-scene contacts violate the assumptions of the raycast module, hindering the global pose  optimisation. One future direction could be accounting for elastic properties of different objects and the integration of more fine-grained deformation models enabled by segmenting a single object into rigid and non-rigid parts. Next, modelling interactions between subjects wearing loose clothes and a non-rigid environment is an intriguing direction.

\noindent\textbf{Closing Remarks.} 
We present \textit{MoCapDeform}, the first framework for markerless global 3D human motion capture from monocular RGB images with the awareness of non-rigid scene deformations. Benefiting from our new raycast-based contact localisation and joint scene deformation and pose optimisation steps, we find accurate global human poses and, at the same time, reasonable scene deformations. We show significantly improved global 3D human poses compared to several competing approaches. Due to these encouraging results, we expect in future to see more research on human motion capture systems that are aware of scene changes, including non-rigid deformations. 

\noindent\textbf{Acknowledgements.} This work was supported by the ERC Consolidator Grant 4DRepLy (770784). 

{\small
\bibliographystyle{ieee_fullname}
\bibliography{egbib}
}

\end{document}